\documentclass[preprint,12pt,number]{elsarticle}
\usepackage{amssymb}
\usepackage{amsmath}
\usepackage{pdfpages}

\usepackage[colorlinks=true, linkcolor=blue, citecolor=blue, urlcolor=blue]{hyperref}

\journal{Pattern Recognition}

\begin{document}

\begin{frontmatter}



\title{Adaptive Visual Conditioning for Semantic Consistency in Diffusion-Based Story Continuation} %% Article title


\author[a]{Seyed Mohammad Mousavi} %% Author name
\author[a]{Morteza Analoui}
\affiliation[a]{organization={School of Computer Engineering, Iran University of Science and Technology},%Department and Organization
            country={Iran}}

%% Abstract
\begin{abstract}
Story continuation focuses on generating the next image in a narrative sequence so that it remains coherent with both the ongoing text description and the previously observed images. A central challenge in this setting lies in utilizing prior visual context effectively, while ensuring semantic alignment with the current textual input. In this work, we introduce \textbf{AVC} (Adaptive Visual Conditioning), a framework for diffusion-based story continuation. AVC employs the CLIP model to retrieve the most semantically aligned image from previous frames. Crucially, when no sufficiently relevant image is found, AVC adaptively restricts the influence of prior visuals to only the early stages of the diffusion process. This enables the model to exploit visual context when beneficial, while avoiding the injection of misleading or irrelevant information. Furthermore, we improve data quality by re-captioning a noisy dataset using large language models, thereby strengthening textual supervision and semantic alignment.  Quantitative results and human evaluations demonstrate that AVC achieves superior coherence, semantic consistency, and visual fidelity compared to strong baselines, particularly in challenging cases where prior visuals conflict with the current input.
\end{abstract}


% \begin{highlights}
% \end{highlights}

%% Keywords

\end{frontmatter}



\section{Introduction}
Diffusion-based models such as DALL·E 2~\cite{ramesh2022hierarchical}, Imagen~\cite{saharia2022photorealistic}, and Stable Diffusion~\cite{rombach2022high} have achieved remarkable success in text-to-image generation, producing visually coherent and semantically accurate results from textual prompts. However, when applied to story continuation—the task of generating the next image conditioned on the current textual description and prior images—their performance often falls short in consistency, narrative flow, and context preservation.

Unlike standalone text-to-image tasks, story continuation inherently requires temporal and visual coherence; each frame must not only reflect the current sentence but also maintain meaningful continuity with prior frames. This dual conditioning introduces challenges such as preserving backgrounds, handling scene transitions, and managing character continuity or intentional forgetting when contexts change. Existing models (e.g., AR-LDM~\cite{pan2024synthesizing}) often perform well only on curated datasets and lack generalization beyond them.

Recent work, such as StoryGen~\cite{liu2024intelligent}, attempted to address generalization by training on the large-scale StorySalon dataset in a zero-shot setting. While this improved generalizability, the treatment of visual memory remained limited. Specifically, models often incorporate previous images statically without evaluating their relevance to the current context, which can lead to visual artifacts and semantic drift.

In this paper, we introduce \textbf{AVC (Adaptive Visual Conditioning)}, an inference-time strategy for story continuation that dynamically adjusts the influence of prior visuals according to their semantic alignment with the current textual input. Our contributions are as follows:

\begin{itemize}
\item Data Re-captioning for Enhanced Alignment: We improve visual-textual alignment by re-captioning weakly annotated datasets using large language models (e.g., GPT~\cite{achiam2023gpt}), producing more descriptive and consistent annotations.
\item CLIP-Based Semantic Memory Selection: We employ CLIP~\cite{radford2021learning} to rank previous images by similarity to the current sentence, selecting only the most relevant frame as visual memory.
\item Adaptive Conditioning Mechanism: When no sufficiently relevant frame is available, AVC reduces the influence of visual memory by restricting its effect to early diffusion timesteps, preventing semantic drift and preserving coherence.
\end{itemize}
\section{Related Works}
% \label{sec1}
%% Labels are used to cross-reference an item using \ref command.

% Section text. See Subsection \ref{subsec1}.

%% Use \subsection commands to start a subsection.
\subsection{Diffusion Models}
% \label{subsec1}
Diffusion models~\cite{sohl2015deep} have recently emerged as a powerful class of generative models. Their core idea is to gradually add noise to training data and then learn to reverse this process to recover the original data distribution. DDPMs~\cite{ho2020denoising} train a sequence of probabilistic models to reverse each noise step, using analytical approximations of the reverse process for efficient training. SMLDs~\cite{song2019generative, song2020improved} estimate the gradient of the data log-density (the score) at various noise levels and use Langevin dynamics to denoise samples. For faster inference, models such as DDIM~\cite{song2020denoising} reduce the number of denoising steps while maintaining sample quality. \\
Building upon these methods, diffusion models have demonstrated remarkable success across a wide range of applications, including inpainting~\cite{lugmayr2022repaint,meng2021sdedit,saharia2022palette}, super-resolution~\cite{saharia2022image,saharia2022palette,rombach2022high}, and conditional generation~\cite{dhariwal2021diffusion,ramesh2022hierarchical,rombach2022high,saharia2022photorealistic}.

\subsection{Text-to-Image Generation}
The goal of text-to-image generation is to synthesize images aligned with natural language prompts. Early approaches were dominated by GAN-based~\cite{goodfellow2020generative} models such as StackGAN~\cite{zhang2017stackgan} and AttnGAN~\cite{xu2018attngan}, which introduced hierarchical generation and attention mechanisms to improve semantic alignment. These models are trained through an adversarial process, in which a generator network learns to synthesize data samples while a discriminator network simultaneously learns to distinguish real samples from those generated by the generator. Later, auto-regressive approaches like VQ-VAE~\cite{van2017neural} and DALL·E~\cite{ramesh2021zero} enabled discrete latent representations. These models factorize the joint distribution into a sequence of conditional probabilities. \\
Diffusion-based text-to-image models, such as Imagen~\cite{saharia2022photorealistic} and DALL-E 2~\cite{ramesh2022hierarchical}, are widely adopted for their impressive generative capabilities. Among these, Stable Diffusion~\cite{rombach2022high} is especially popular and operates in latent space. This approach increases efficiency while maintaining high-quality image production. Stable Diffusion is often used as a baseline in research.

\subsection{Story Synthesis}
Story visualization aims to generate a sequence of coherent images that correspond to a multi-sentence narrative, making it more complex than single-turn text-to-image synthesis. A related variant, known as story continuation, shares the same goal but further conditions generation on a given source frame. Early works in story visualization, such as StoryGAN~\cite{li2019storygan} on the PororoSV dataset, introduced GAN-based architectures to capture temporal relationships across story sequences. Later methods, such as DUCO-StoryGAN~\cite{maharana2021improving}, enhanced this direction through dual learning and copy-transform. In contrast, story continuation was explored in StoryDALL·E~\cite{maharana2022storydall}, which leveraged a pre-trained DALL·E~\cite{ramesh2021zero} model for narrative-driven image synthesis. Diffusion-based approaches have recently advanced the field: AR-LDM~\cite{pan2024synthesizing} introduced autoregressive conditioning within a latent diffusion framework to improve consistency across frames. Building on this line, ACM-VSG~\cite{feng2023improved} proposed adaptive context modeling to strengthen visual memory and temporal consistency throughout a story. Most recently, StoryGen~\cite{liu2024intelligent} addressed the challenge in a zero-shot setting by training on a large-scale dataset (StorSalon); however, it still suffered from a limited modeling of visual memory, often failing to adapt when previous frames were only weakly relevant to the current text.
%% Use \subsubsection, \paragraph, \subparagraph commands to 
%% start 3rd, 4th and 5th level sections.
%% Refer following link for more details.
%% https://en.wikibooks.org/wiki/LaTeX/Document_Structure#Sectioning_commands

% \subsubsection{Mathematics}
%% Inline mathematics is tagged between $ symbols.
% This is an example for the symbol $\alpha$ tagged as inline mathematics.

%% Displayed equations can be tagged using various environments. 
%% Single line equations can be tagged using the equation environment.

\section{Method}

\subsection{Problem Formulation}
We formulate story continuation as follows. Given a sequence of text–image
pairs
\[
\{(s_1, I_1), (s_2, I_2), \ldots, (s_{t-1}, I_{t-1})\}
\]
together with the current sentence $s_t$, the objective is to generate the
next image $I_t$ that is semantically consistent with $s_t$ while preserving
coherence with the preceding narrative.

\paragraph{Diffusion Model}  
Our generative backbone is Stable Diffusion~\cite{rombach2022high}, a latent denoising diffusion model based on DDPMs~\cite{ho2020denoising}. The forward (noising) process gradually corrupts a clean image
$I_0$ into a latent variable $x_t$ through Gaussian perturbations:
\[
q(x_t \mid x_{t-1}) = \mathcal{N}\bigl(x_t; \sqrt{\alpha_t}\, x_{t-1}, (1 - \alpha_t)\mathbf{I}\bigr),
\]
where $\alpha_t = 1 - \beta_t$ and $\beta_t \in (0,1)$ is a variance
schedule.  
Defining the cumulative product
\[
\overline{\alpha}_t = \prod_{i=1}^t \alpha_i,
\]
we can directly express the noised sample at step $t$ as
\[
q(x_t \mid x_0) = \mathcal{N}\bigl(x_t; \sqrt{\overline{\alpha}_t}\, x_0, (1 - \overline{\alpha}_t)\mathbf{I}\bigr).
\]

\paragraph{Reverse Process}  
The reverse denoising process is parameterized by a UNet~\cite{ronneberger2015u}, which predicts the
added noise $\epsilon_\theta(x_t, t)$. The generative distribution is defined as
\[
p_\theta(x_{t-1} \mid x_t) = \mathcal{N}\Bigl(x_{t-1}; 
\mu_\theta(x_t, t), \sigma_t^2 \mathbf{I}\Bigr),
\]
where the mean is computed as
\[
\mu_\theta(x_t, t) =
\frac{1}{\sqrt{\alpha_t}} \left(x_t - \frac{1 - \alpha_t}{\sqrt{1 - \overline{\alpha}_t}}\, \epsilon_\theta(x_t, t)\right).
\]

\paragraph{Classifier-Free Guidance}  
To enhance semantic control, we employ classifier-free guidance~\cite{ho2022classifier}. The denoising
UNet is trained with both conditional and unconditional inputs, enabling
guided sampling at inference:
\[
\epsilon_\theta^{\text{guid}} = (1 + w)\,\epsilon_\theta(x_t, t, c) - w\,\epsilon_\theta(x_t, t, \varnothing),
\]
where $w$ is the guidance scale.
In our setting, the conditioning $c$ includes the current text $x_t$ and
adaptively selected prior visual context.
Following StoryGen~\cite{liu2024intelligent}, which builds upon the conditional formulation introduced in InstructPix2Pix~\cite{brooks2023instructpix2pix}, the noise prediction network is extended to incorporate both image and text conditions. Specifically, the denoising function is modified as
\begin{equation*}
	\begin{aligned}
		\tilde{\epsilon}_{\theta}(z_t, c_I, c_T) &= \epsilon_{\theta}(z_t, \varnothing, \varnothing) \\
		&\quad + s_I \cdot \big(\epsilon_{\theta}(z_t, c_I, \varnothing) - \epsilon_{\theta}(z_t, \varnothing, \varnothing)\big) \\
		&\quad + s_T \cdot \big(\epsilon_{\theta}(z_t, c_I, c_T) - \epsilon_{\theta}(z_t, c_I, \varnothing)\big),
	\end{aligned}
\end{equation*}
where $c_I$ and $c_T$ denote the image and text conditions, respectively, while $s_I$ and $s_T$ are guidance scales controlling the contribution of each modality. This formulation generalizes classifier-free guidance to the multimodal setting.

\subsection{Data Re-captioning}

A major challenge in story continuation lies in the low quality and inconsistency of textual annotations accompanying images. Many captions of the StorySalon dataset~\cite{liu2024intelligent} are either incomplete or semantically misaligned with the associated images. Such inconsistencies have a severe impact on downstream tasks, particularly the selection of semantically relevant prior frames for conditioning, as misaligned captions hinder accurate comparisons of similarity between the current textual input and previous visual content. 

Given that one of the core components of our method is retrieving the most semantically aligned past frame with respect to the current narrative, it is essential to first improve the textual quality of image descriptions.

Specifically, we employed three captioning strategies:

\begin{enumerate}
	\item BLIP~\cite{li2022blip} — a vision-language model trained on large-scale image and text data. It was a groundbreaking model for generating descriptive image captions.
	\item BLIP-2~\cite{li2023blip} — an advancement that connects a frozen pre-trained image encoder with a frozen pre-trained large language model (LLM) using a new, lightweight module called the Querying Transformer (Q-Former). This enables it to leverage the powerful capabilities of LLMs for more complex, conversational, and contextually rich captions without requiring extensive training.
	\item GPT-based multimodal captioning~\cite{achiam2023gpt} — a large language model prompted with the image to generate fluent, semantically rich descriptions with higher linguistic naturalness.
\end{enumerate}

To evaluate the quality of the re-captioned StorySalon test set, we computed CLIP similarity scores between each generated caption and its corresponding image. Let $s_{i,j}$ denote the CLIP cosine similarity for image $i$ and caption from model $j \in \{\text{BLIP, BLIP-2, GPT}\}$. We then report the mean and standard deviation of these scores across the test set:
\begin{equation*}
	\mu_j = \frac{1}{N} \sum_{i=1}^{N} s_{i,j}, \quad
	\sigma_j = \sqrt{\frac{1}{N} \sum_{i=1}^{N} (s_{i,j} - \mu_j)^2}.
\end{equation*}

\begin{figure}
	\centering
	\includegraphics[width=0.45\linewidth]{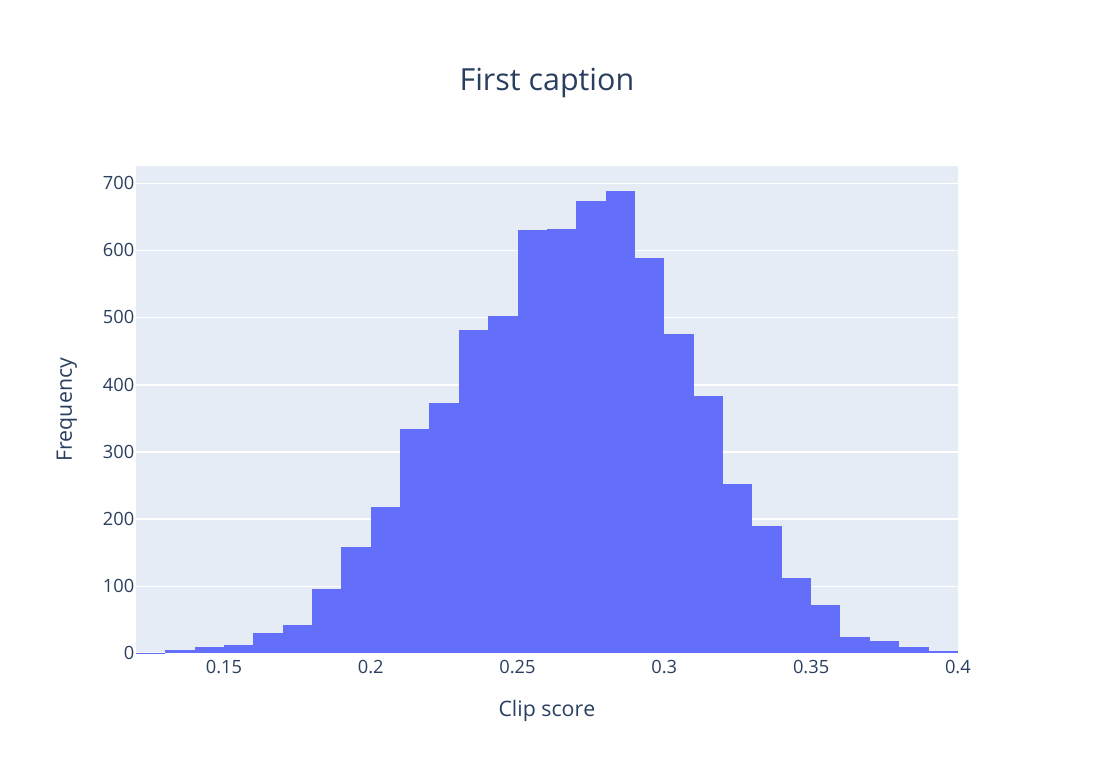}
	\includegraphics[width=0.45\linewidth]{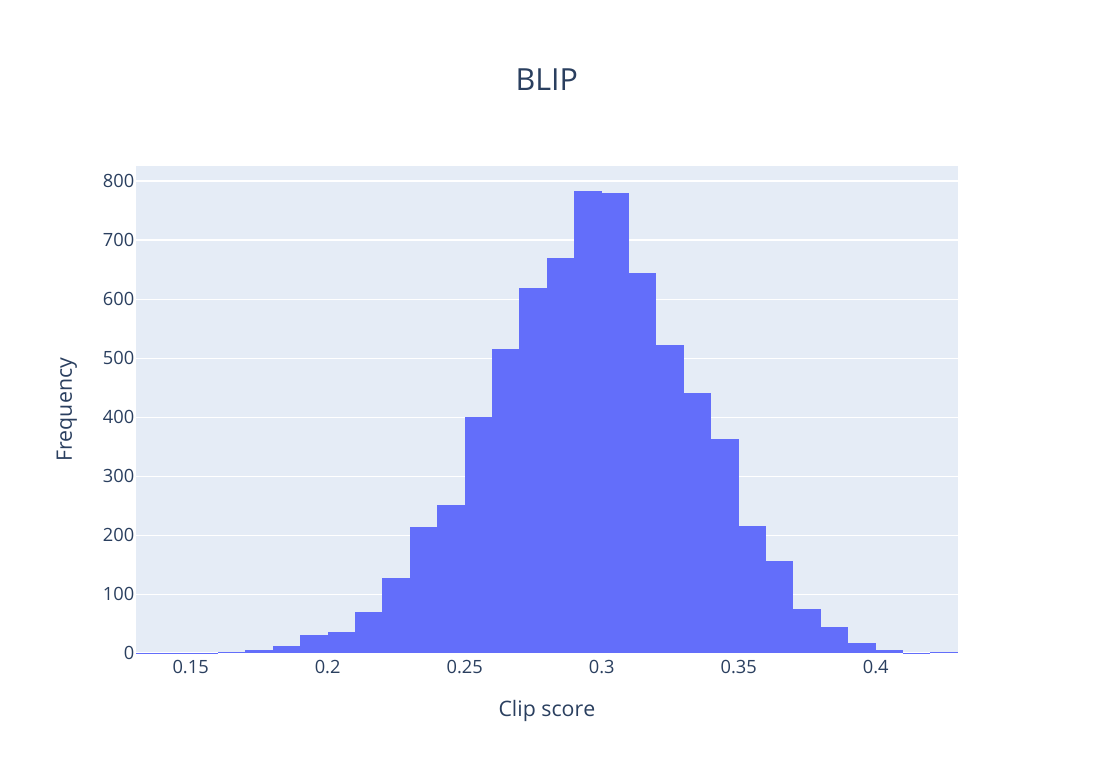}\\
	\includegraphics[width=0.45\linewidth]{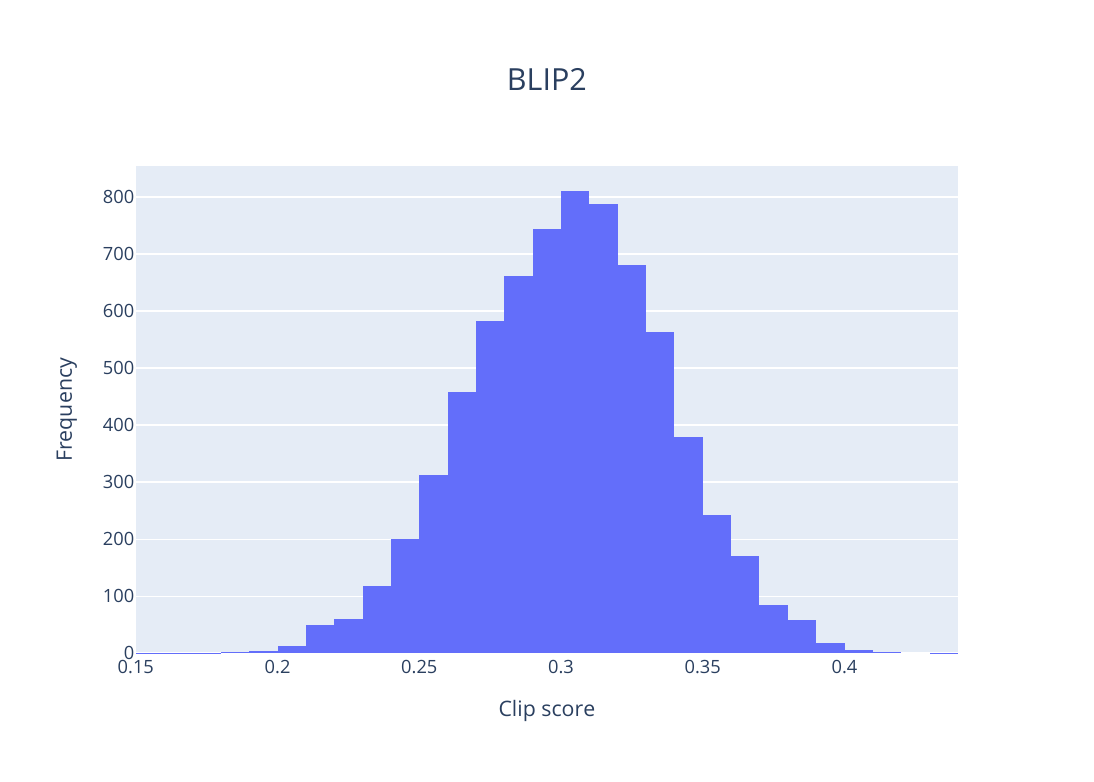}
	\includegraphics[width=0.45\linewidth]{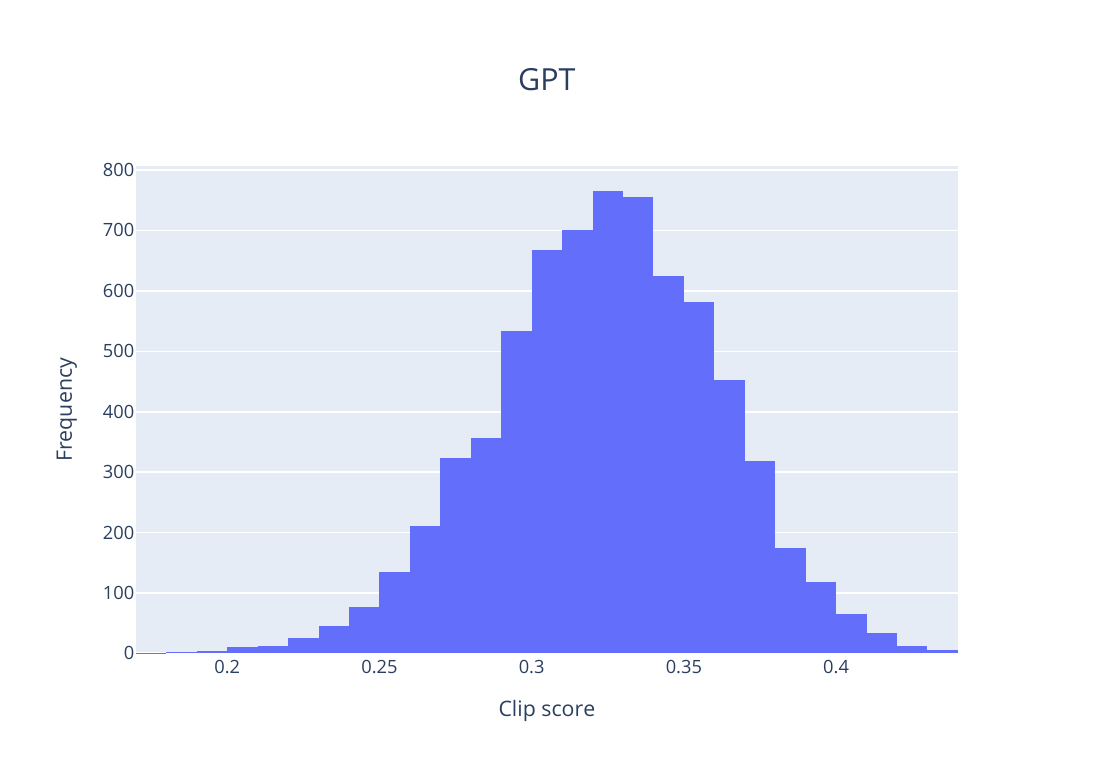}
	\caption{Distributions of CLIP similarity scores for first caption, BLIP, BLIP-2 and GPT captions on the StorySalon test set.}
	\label{fig:caption_distributions}
\end{figure}

Figure~\ref{fig:caption_distributions} visualizes the distributions of CLIP scores for each model, while Table~\ref{tab:clip_scores} summarizes the mean and standard deviation. As shown, GPT-based captions achieve the highest semantic alignment with images, confirming that large language models provide superior textual grounding for subsequent semantic image selection in the AVC framework.

\begin{table}[t]
	\centering
	\begin{tabular}{lcc}
		\hline
		Model & Avg & Std \\
		\hline
		First caption & 0.27 & 0.04 \\
		BLIP  & 0.30 & 0.04 \\
		BLIP-2 & 0.30 & 0.03 \\
		\textbf{GPT}   & \textbf{0.32} & \textbf{0.04} \\
		\hline
	\end{tabular}
	\caption{Mean and standard deviation of CLIP similarity scores for re-captioned StorySalon test set. Higher is better.}
	\label{tab:clip_scores}
\end{table}

\subsection{CLIP-based Semantic Image Selection}

A key challenge in story continuation is determining which previous frame to leverage as visual context. Not all prior images are equally relevant, and some may introduce inconsistencies if used directly. To address this, we adopt a CLIP-based similarity scoring mechanism that jointly considers both textual and visual alignment.

Formally, given the current textual input $x_t$, we compute two similarity scores:

\begin{itemize}
	\item \textbf{Textual similarity:} For each previous text $x_j$, we calculate
	\[
	s_{\text{text}}(x_t, x_j) = \frac{\langle f_{\text{CLIP}}^{\text{text}}(x_t), f_{\text{CLIP}}^{\text{text}}(x_j) \rangle}{\| f_{\text{CLIP}}^{\text{text}}(x_t)\| \cdot \| f_{\text{CLIP}}^{\text{text}}(x_j)\|},
	\]
	where $f_{\text{CLIP}}^{\text{text}}(\cdot)$ denotes the CLIP text encoder.
	
	\item \textbf{Visual similarity:} For each previous frame $I_j$, we compute
	\[
	s_{\text{image}}(x_t, I_j) = \frac{\langle f_{\text{CLIP}}^{\text{text}}(x_t), f_{\text{CLIP}}^{\text{image}}(I_j) \rangle}{\| f_{\text{CLIP}}^{\text{text}}(x_t)\| \cdot \| f_{\text{CLIP}}^{\text{image}}(I_j)\|},
	\]
	where $f_{\text{CLIP}}^{\text{image}}(\cdot)$ denotes the CLIP image encoder.
\end{itemize}

Since $s_{\text{text}}$ and $s_{\text{image}}$ are not directly comparable in scale, we normalize each score distribution using Z-score normalization:
\[
\tilde{s} = \frac{s - \mu}{\sigma},
\]
where $\mu$ and $\sigma$ are the mean and standard deviation of the similarity scores for the respective modality.

Finally, we compute the average normalized score for each candidate frame:
\[
S_j = \frac{1}{2}\left( \tilde{s}_{\text{text}}(x_t, x_j) + \tilde{s}_{\text{image}}(x_t, I_j) \right).
\]

The selected frame is the one with the highest combined similarity:
\[
I^* = \arg\max_j S_j.
\]

\subsection{Adaptive Visual Conditioning}
\begin{figure*}[t]
	\centering
	\includegraphics[width=\textwidth]{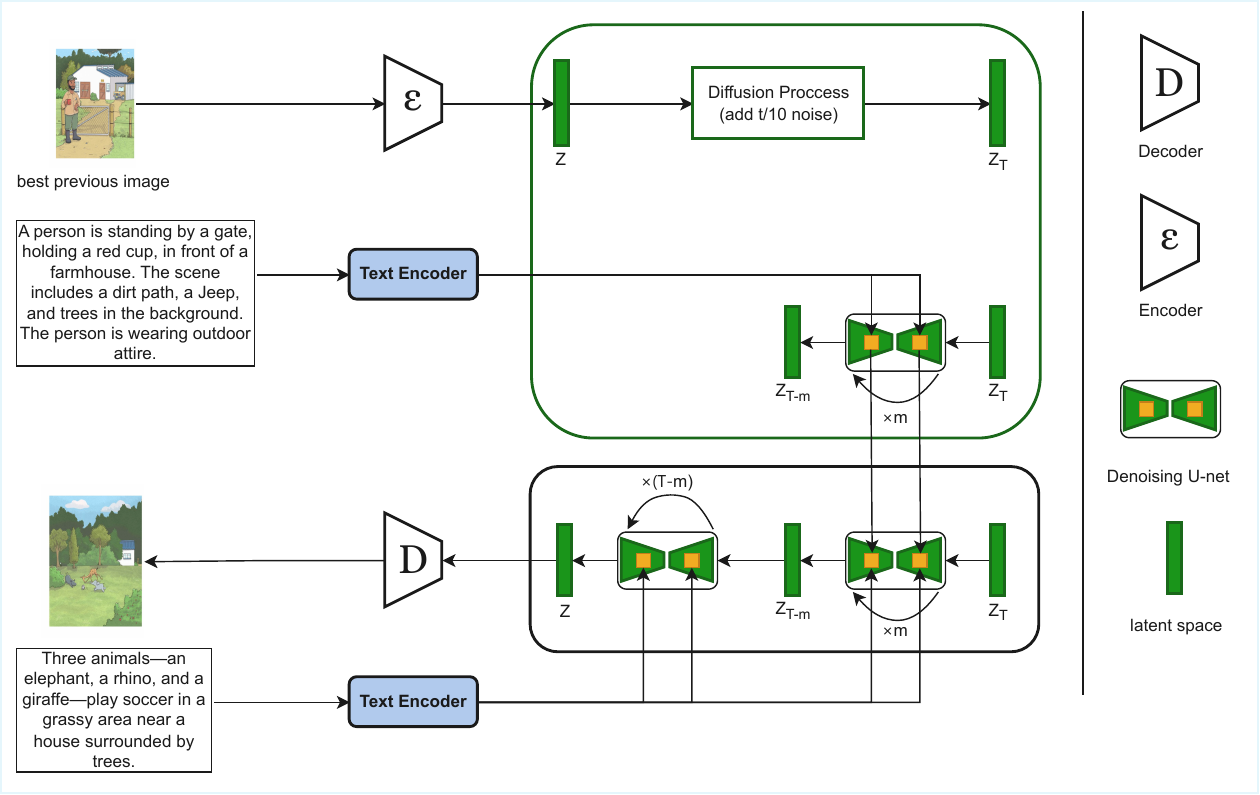}
	\caption{Overview of our proposed Adaptive Visual Conditioning (AVC) framework. 
		Given the current text description and the most semantically relevant previous frame (selected by CLIP-based similarity), both are encoded into the latent space. 
		The image latent is perturbed through the diffusion process, while the text embedding provides semantic guidance. 
		Depending on the similarity score $s$, Image conditioning is injected adaptively up to timestep $m(s)$, after which only the text condition remains: 
		(i) for low $s$, image guidance is applied only in early steps, 
		(ii) for medium $s$, it is gradually extended, and 
		(iii) for high $s$, both text and image are used throughout the full denoising process. 
		This adaptive design balances reliance on textual and visual information based on the reliability of the retrieved frame.}
	\label{fig:avc_strategy}
\end{figure*}

Although CLIP-based Semantic Image Selection helps identify the most relevant previous frame, in some cases, the similarity score remains low, indicating that no earlier frame provides sufficient semantic alignment with the current text. To address this, we adopt an \emph{Adaptive Visual Conditioning} (AVC) strategy, where the influence of the selected image condition is adapted according to its similarity score, as illustrated in Figure~\ref{fig:avc_strategy}.

Formally, given the similarity score $s$ of the selected frame, we define the timestep $m$ at which image conditioning is injected into the diffusion process as:
\begin{equation*}
	m(s) = 
	\begin{cases}
		m_{\min}, & s \leq \tau_{\min}, \\[6pt]
		\left\lfloor m_{\min} + \dfrac{(s - \tau_{\min})(T - m_{\min})}{\tau_{\max} - \tau_{\min}} \right\rfloor, & \tau_{\min} < s < \tau_{\max}, \\[12pt]
		T, & s \geq \tau_{\max},
	\end{cases}
\end{equation*}
where $T$ is the total number of diffusion steps, $m_{\min}$ is the minimum timestep for applying dual conditioning (text and image), $\tau_{\min}$ is the minimum similarity threshold, and $\tau_{\max}$ is the maximum similarity threshold.  

Intuitively, when the score is low ($s \leq \tau_{\min}$), AVC reduces the model's reliance on the image by injecting it only in the earliest timesteps, allowing the model to rely primarily on text. As the score increases, the conditioning is extended to later steps, and when $s \geq \tau_{\max}$, both text and image conditions are applied throughout all timesteps. This adaptive mechanism ensures a balanced integration of textual and visual guidance depending on the quality of the retrieved frame.

\section{Experiments}

\subsection{Experimental Settings}
Our framework is built directly on the pre-trained weights of \textit{StoryGen}~\cite{liu2024intelligent} and is evaluated on the \textit{StorySalon} dataset introduced in the same work. For evaluation, we follow the official test split of StorySalon, which contains $7{,}018$ image--text pairs organized into $515$ folders, where each folder corresponds to a specific story. Since our contributions focus exclusively on inference-time strategies, no additional training is performed. To ensure a fair comparison with the baseline, we adopt the same hyperparameter settings: the classifier-free guidance scales are set to $s_I = 7.0$ for image conditions and $s_T = 3.5$ for text conditions. The total number of diffusion timesteps is fixed to $40$, and for conditional diffusion, the image condition is injected up to $t' = t/10$. During inference, only one previous (reference) image is used, consistent with the StoryGen setup.  

All experiments are implemented in PyTorch and conducted on a single NVIDIA RTX~3090 GPU with a batch size of $1$ at a resolution of $512 \times 512$. Random seeds are fixed for reproducibility.  Inference time depends on the number of timesteps and the use of dual conditions: for both image and text conditions, generating a single image takes approximately 45 seconds over 40 timesteps. However, for the less image-conditioned denoise process (i.e., the smaller m), the time will be shorter.

For evaluation, we employ three metrics: CLIP-I, CLIP-T, and FID~\cite{heusel2017gans}. Following StoryGen, we use \textit{PickScore}~\cite{kirstain2023pick} to automatically select the generated images with higher quality. Specifically, each reported score corresponds to the best image chosen from a pool of $10$ candidates.

\subsection{Quantitative Results}
\subsubsection{Re-captioning Performance}
\begin{figure}
	\centering
	\includegraphics[width=\linewidth]{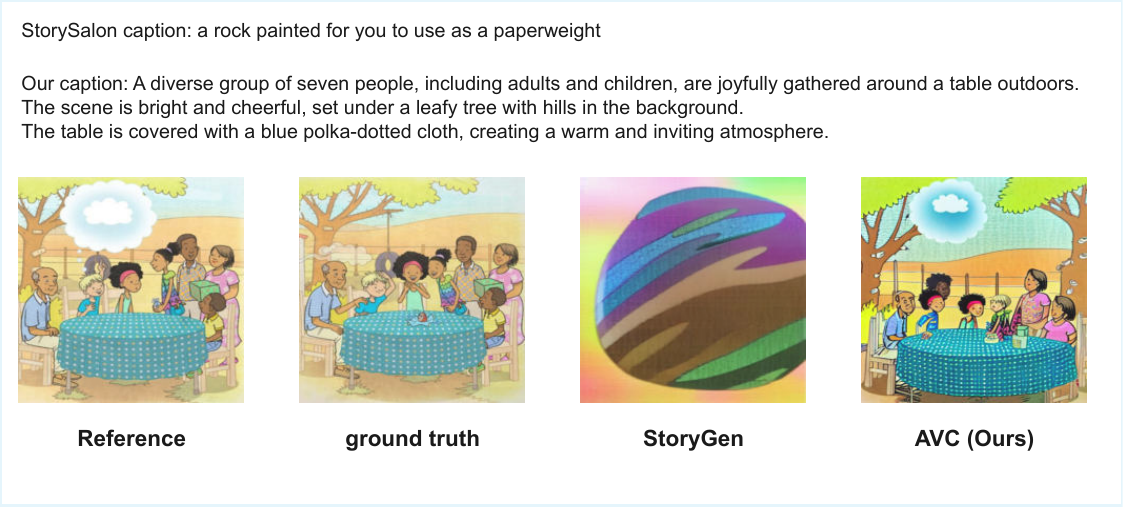}
	\includegraphics[width=\linewidth]{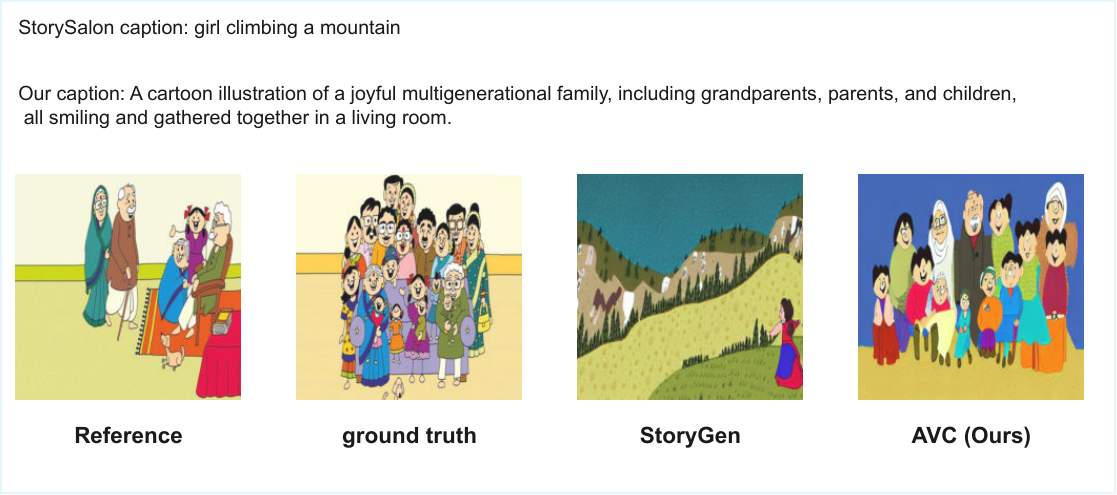}\\
	\caption{Qualitative examples of re-captioning. The new captions lead to semantically richer descriptions and more visually coherent generations compared to the previous captions.}
	\label{fig:recaptioning-examples}
\end{figure}

We first evaluate the effect of re-captioning on $1{,}030$ samples (two frames per story). 
As shown in Table~\ref{tab:recaptioning}, our method achieves clear improvements over the original captions in all metrics. 
Figure~\ref{fig:recaptioning-examples} further illustrates how re-captioning produces semantically richer descriptions, which in turn lead to visually more coherent generations.

\begin{table}[t]
	\centering
	\begin{tabular}{lccc}
		\hline
		Method & CLIP-I $\uparrow$ & CLIP-T $\uparrow$ & FID $\downarrow$ \\
		\hline
		Prev. Captions & 0.7449 & 0.2694 & 74.11 \\
		New Captions   & \textbf{0.7721} & \textbf{0.3318} & \textbf{73.97} \\
		\hline
	\end{tabular}
	\caption{Re-captioning results on 1,030 samples.}
	\label{tab:recaptioning}
\end{table}

\subsubsection{Effect of CLIP-based Selection}
\begin{figure}
	\centering
	\includegraphics[width=\linewidth]{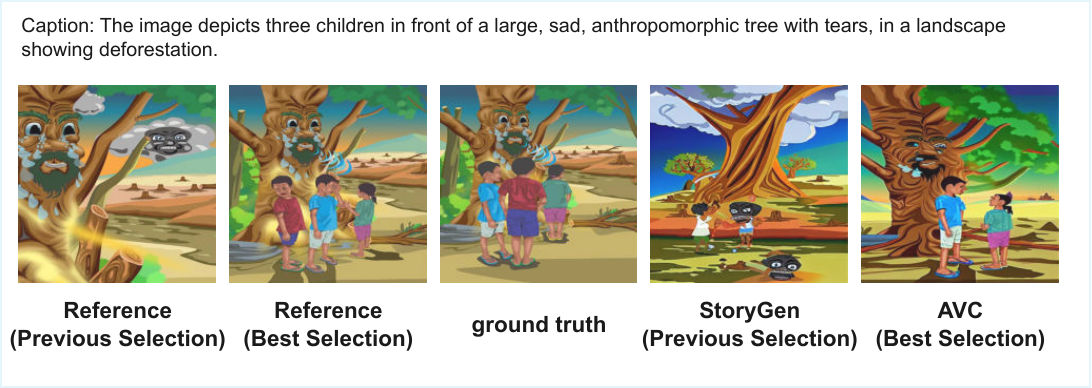}
	\includegraphics[width=\linewidth]{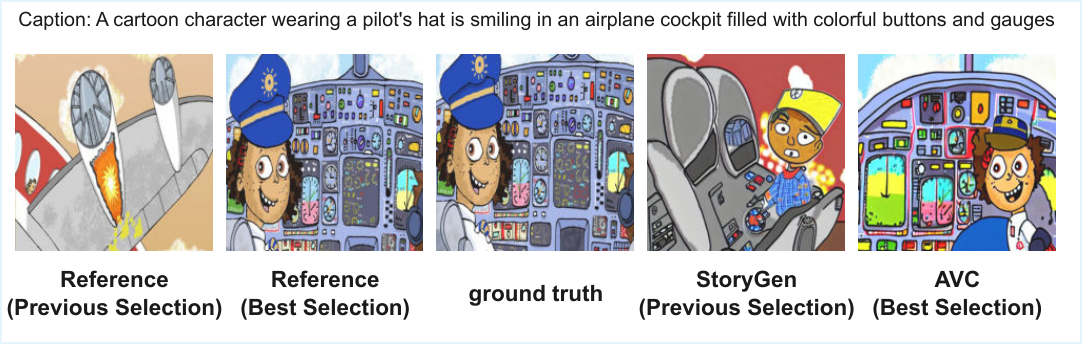}\\
	\caption{Examples of CLIP-based best image selection. Compared to the previous selection, the chosen images better align with the textual descriptions under new captions.}
	\label{fig:clip-selection}
\end{figure}

Next, we evaluate the impact of our CLIP-based best image selection strategy. 
At this stage, \textit{re-captioning} has already been applied, ensuring that the comparison isolates the effect of the CLIP-based selection itself. 
In other words, we only assess the contribution of the proposed CLIP-based selection strategy. Unlike the baseline StoryGen, which always uses the last image as the reference, our method identifies and selects the best image according to CLIP similarity.
As expected, improvements are larger for subsets with a higher \textit{score difference}, i.e., where the initial selection was suboptimal. 
Table~\ref{tab:clip-selection-combined} reports quantitative results, and qualitative examples are shown in Figure~\ref{fig:clip-selection}.

\begin{table*}[!t]
	\centering
	\begin{tabular}{lcccc}
		\hline
		Subset & Method & CLIP-I $\uparrow$ & CLIP-T $\uparrow$ & FID $\downarrow$ \\
		\hline
		500 high-\textit{score difference}  & Prev. Selection & 0.7635 & 0.3272 & 107.78 \\
		& Best Selection  & \textbf{0.7806} & \textbf{0.3390} & \textbf{105.28} \\
		\hline
		1,987 high-\textit{score difference} & Prev. Selection & 0.7618 & 0.3265 & 52.39 \\
		& Best Selection  & \textbf{0.7744} & \textbf{0.3352} & \textbf{50.94} \\
		\hline
	\end{tabular}
	\caption{Comparison of CLIP-based selection on two subsets of samples with the highest \textit{score difference}. Best Selection consistently outperforms the previous strategy.}
	\label{tab:clip-selection-combined}
\end{table*}

\subsubsection{Adaptive Visual Conditioning (AVC)}
\begin{figure}
	\centering
	\includegraphics[width=\linewidth]{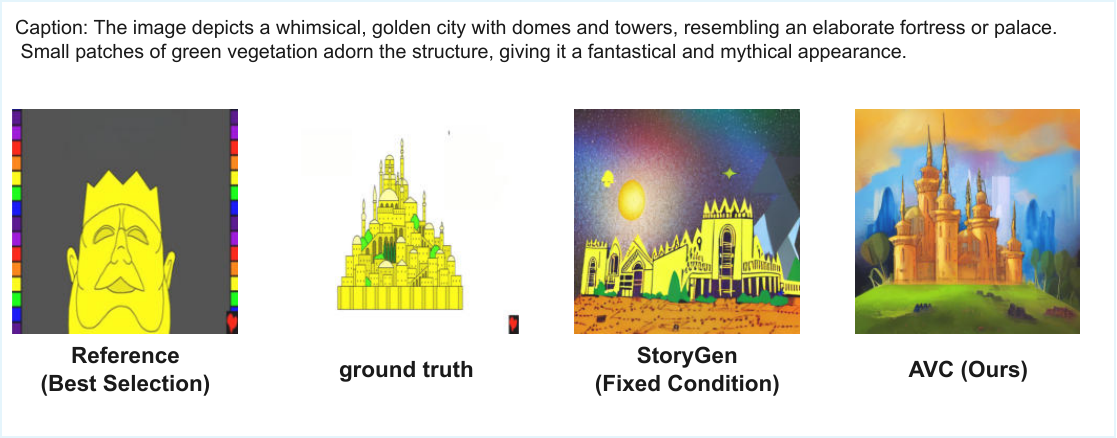}
	\includegraphics[width=\linewidth]{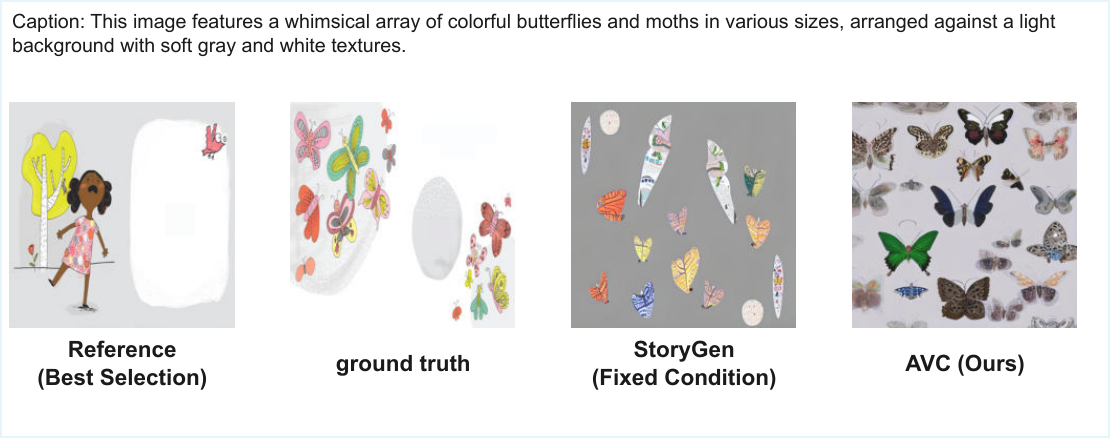}\\
	\caption{Qualitative examples of Adaptive Visual Conditioning (AVC). AVC enhances alignment between captions and generated images, with improved text consistency and reduced visual artifacts compared to fixed-timestep conditioning.}
	\label{fig:avc-examples}
\end{figure}

In this setting, the focus is on cases where the similarity score between the current prompt and the previous frames is relatively low, which poses a challenge for stable conditioning. 
To this end, we select $3{,}011$ samples that meet this criterion. 
At this stage, \textit{re-captioning} and \textit{CLIP-based data selection} have already been applied, ensuring that the comparison isolates the contribution of AVC itself. 
In other words, we only assess the effect of the proposed AVC strategy under these challenging conditions. 
Results are reported in Table~\ref{tab:avc}. 
AVC consistently improves CLIP-T and FID, demonstrating stronger text–image alignment and improved image quality. 
Although CLIP-I remains close to the fixed-timestep case (0.7608 vs.~0.7618), the overall trend confirms the effectiveness of AVC. 
Figure~\ref{fig:avc-examples} highlights qualitative improvements.

\begin{table}[t]
	\centering
	\begin{tabular}{lccc}
		\hline
		Method & CLIP-I $\uparrow$ & CLIP-T $\uparrow$ & FID $\downarrow$ \\
		\hline
		Fixed condition & \textbf{0.7618} & 0.3208 & 43.78 \\
		AVC     & 0.7608 & \textbf{0.3277} & \textbf{41.2} \\
		\hline
	\end{tabular}
	\caption{Results for Adaptive Visual Conditioning (AVC) on 3,011 samples.}
	\label{tab:avc}
\end{table}

\subsubsection{Overall Performance}

This section provides the final, comprehensive comparison between our AVC framework and the leading state-of-the-art baselines across two distinct data conditions: using the original, noisy captions and using our refined, high-quality recaptions.

\paragraph{Performance with Original Captions}
Table \ref{tab:original_caption_performance} presents the baseline performance using the original captions on the full StorySalon test set. The results show that our AVC model substantially reduces FID to 30.14, representing a notable improvement in perceptual quality, while maintaining comparable CLIP-I and CLIP-T scores (0.7438 and 0.2835, respectively). 
This suggests that AVC effectively enhances image realism and reduces generation artifacts without compromising semantic alignment with either the text or ground-truth images. 

\begin{table}[t]
	\centering
	\begin{tabular}{lccc}
		\hline
		Model & CLIP-I $\uparrow$ & CLIP-T $\uparrow$ & FID $\downarrow$ \\
		\hline
		StoryDALL$\cdot$E & 38.34 & 0.6823 & 0.2366 \\
		AR-LDM & 39.55 & 0.6864 & 0.2614 \\
		StoryGen & 33.90 & \textbf{0.7467} & \textbf{0.2875} \\
		AVC (ours) & \textbf{30.14} & 0.7438 & 0.2835 \\
		\hline
	\end{tabular}
	\caption{Performance with Original Caption (Full Dataset).}
	\label{tab:original_caption_performance}
\end{table}

\paragraph{Performance with Recaptioning (Final Comparison)}
Table~\ref{tab:recaption_performance_full} presents the final comparison using the improved LLM-generated re-captions applied to the full dataset, representing the optimal configuration of our framework. 
The proposed AVC method achieves state-of-the-art performance across all evaluation metrics, demonstrating the effectiveness of integrating high-quality semantic guidance from re-captioning with the adaptive visual conditioning mechanism. 
These results confirm that enhancing textual precision and dynamically adjusting visual conditioning jointly contribute to superior semantic alignment and perceptual fidelity.

\begin{table}[t]
	\centering
	\begin{tabular}{lccc}
		\hline
		Model & CLIP-I $\uparrow$ & CLIP-T $\uparrow$ & FID $\downarrow$ \\
		\hline
		StoryGen & 32.41 & 0.7710 & 0.3294 \\
		AVC (ours) & \textbf{30.86} & \textbf{0.7752} & \textbf{0.3361} \\
		\hline
	\end{tabular}
	\caption{Performance with Recaption (Full Dataset). AVC achieves the best results across all metrics, establishing state-of-the-art performance.}
	\label{tab:recaption_performance_full}
\end{table}

\subsection{Human Evaluation}
To complement the quantitative metrics, we conducted a comprehensive human evaluation to assess the perceptual quality of the generated images. The evaluation set consisted of 200 images, each rated independently by five evaluators, resulting in a total of 20 raters across the study. All raters were Master's or Ph.D. students in Artificial Intelligence, belonging to the same statistical population to ensure consistency and domain expertise.

Each evaluator rated the images according to three criteria:
\begin{itemize}
	\item \textbf{Semantic Alignment:} The extent to which the generated image visually matches the story caption.
	\item \textbf{Ground-Truth Consistency:} The degree to which the generated image resembles the ground-truth scene.
	\item \textbf{Visual Quality:} The visual appeal of the image and the absence of noticeable artifacts.
\end{itemize}

All ratings were given on a 5-point Likert scale, where 1 indicates the lowest quality and 5 indicates the highest quality.  
The final scores for each criterion were computed by averaging the ratings across all evaluators.  
A summary of the human evaluation results is provided in Table~\ref{tab:human_evaluation}.

\begin{table}[h!]
	\centering
	\begin{tabular}{lccc}
		\hline
		Method & Semantic Alignment & GT Consistency & Visual Quality \\
		\hline
		StoryGen & 2.6370 & 2.3810 & 2.6660 \\
		AVC (ours) & \textbf{2.7850} & \textbf{2.5770} & \textbf{2.7630} \\
		\hline
	\end{tabular}
	\caption{Human evaluation results based on average scores across 20 evaluators (1 = worst, 5 = best).}
	\label{tab:human_evaluation}
\end{table}

\subsection{Ablation Study}
We perform ablation experiments to better understand the contribution of each component in our framework. Specifically, we analyze (i) different strategies for CLIP-based selection, (ii) thresholding parameters $\tau_{\min}, \tau_{\max}$, and (iii) adaptive strategies for timesteps and guidance scale.

\subsubsection{CLIP-Based Selection Strategies}
We evaluate three different approaches for selecting the best image: using only visual similarity (CLIP-I), only textual similarity (CLIP-T), and a combined similarity. We conduct the experiments on $1238$ samples with the highest score difference. Results are summarized in Table~\ref{tab:clip-selection-ablation}.

\begin{table*}[!t]
	\centering
	\begin{tabular}{lccc}
		\hline
		Method & CLIP-I $\uparrow$ & CLIP-T $\uparrow$ & FID $\downarrow$ \\
		\hline
		Last Previous (Paper Idea) & 0.7600 & 0.3235 & 68.22 \\
		Best Previous (Image Only) & 0.7759 & 0.3303 & 66.60 \\
		Best Previous (Text Only)  & 0.7761 & 0.3304 & \textbf{66.05} \\
		Best Previous (Combine)    & \textbf{0.7785} & \textbf{0.3311} & 66.23 \\
		\hline
	\end{tabular}
	\caption{Ablation study of CLIP-based selection strategies on $1238$ high-SD samples. The combined similarity yields the best overall trade-off across metrics.}
	\label{tab:clip-selection-ablation}
\end{table*}

Qualitatively, we observe that the combined approach strikes a balance between both aspects, resulting in the best overall performance.

\subsubsection{Threshold Sensitivity}
We set $m_{\min} = 10$ across all experiments, while $\tau_{\min}$ and $\tau_{\max}$ depend on the selection strategy. For example, the best values for the combined method are $\tau_{\min} = -0.3$ and $\tau_{\max} = 0.85$. For the visual-only method, we found $\tau_{\min} = 0.24$ and $\tau_{\max} = 0.3$.

\subsubsection{Adaptive Timesteps vs. Adaptive Guidance Scale}
We evaluate adaptive strategies for controlling the conditioning process. 
As shown in Table~\ref{tab:adaptive-strategies}, timestep adaptation proves more effective than guidance scale adaptation. 
Reducing the minimum timestep
for applying dual conditioning ($m_{\min}$) enhances semantic alignment, with the 5-step configuration achieving the highest CLIP-T score (0.3390). 
However, excessive reduction slightly degrades perceptual quality, as reflected in higher FID. 
A moderate setting of 10 timesteps yields the best trade-off, producing the lowest FID (105.24) and competitive CLIP-T performance. 
In contrast, varying the adaptive guidance scale leads to relatively minor improvements, with the best CLIP-I (0.7596) obtained at scale 21. 
These results suggest that timestep adaptation provides a more effective mechanism for balancing alignment and visual fidelity.
It is worth noting that these adaptive strategies were evaluated on a subset of 542 samples 
corresponding to the lowest baseline scores, in order to better examine model behavior in challenging cases.

\begin{table*}[!t]
	\centering
	\begin{tabular}{lccc}
		\hline
		Method & CLIP-I $\uparrow$ & CLIP-T $\uparrow$ & FID $\downarrow$ \\
		\hline
		Adaptive Timesteps (20 steps) & 0.7552 & 0.3159 & 107.38 \\
		Adaptive Timesteps (10 steps) & 0.7554 & 0.3322 & \textbf{105.24} \\
		Adaptive Timesteps (5 steps)  & 0.7453 & \textbf{0.3390} & 106.20 \\
		Adaptive Guidance Scale (15)  & 0.7566 & 0.3264 & 108.04 \\
		Adaptive Guidance Scale (21)  & \textbf{0.7596} & 0.3316 & 107.73 \\
		Adaptive Guidance Scale (30)  & 0.7578 & 0.3362 & 108.81 \\
		\hline
	\end{tabular}
	\caption{Comparison of adaptive timestep and adaptive guidance scale strategies. Timestep adaptation provides more stable and effective improvements.}
	\label{tab:adaptive-strategies}
\end{table*}

\subsubsection{Effect of Exponential Mapping in AVC}

\begin{figure}[t]
	\centering
	\includegraphics[width=0.7\textwidth]{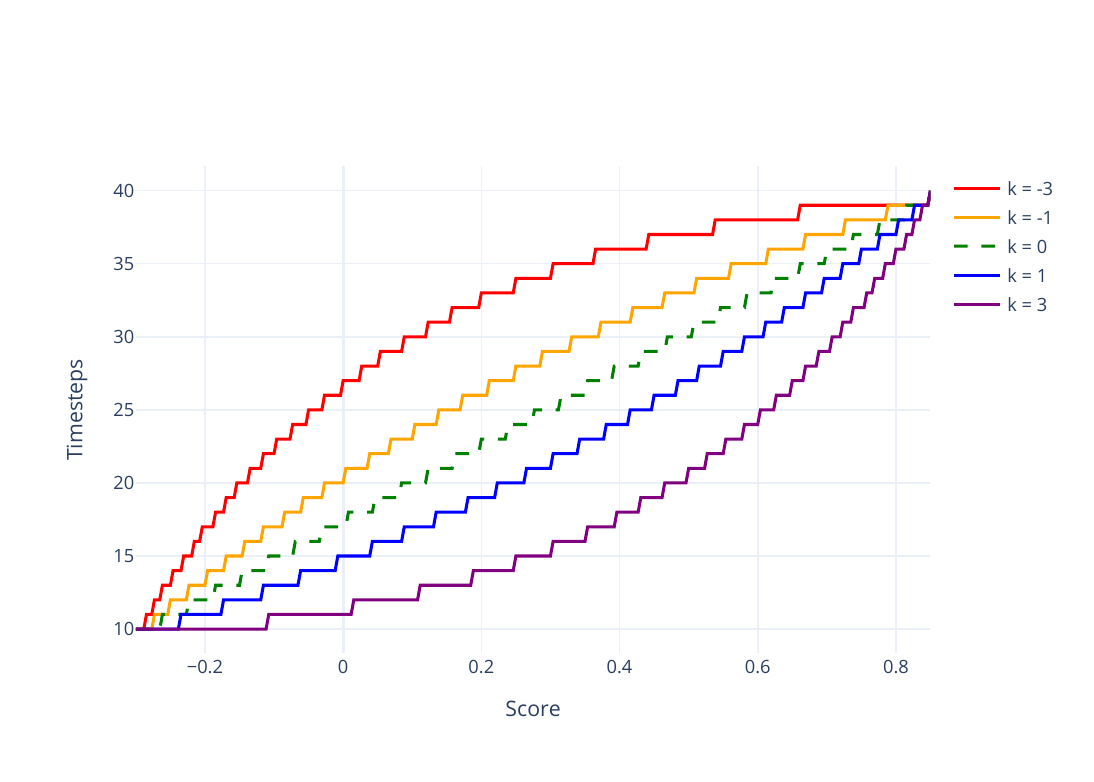}
	\caption{Visualization of timestep allocation for different exponential mapping coefficients ($k \in {-3, -1, 0, 1, 3}$). Positive $k$ values emphasize textual guidance by shortening image-conditioned steps, while negative $k$ values maintain stronger visual conditioning.}
	\label{fig:timesteps_scores}
\end{figure}

To investigate how the scheduling of adaptive conditioning influences image-text alignment and visual fidelity, we conducted experiments on a subset of $3{,}011$ samples with the lowest similarity scores. Specifically, we examined the effect of using exponential mappings for timestep scheduling in AVC. The mapping is defined as
\[
m(s) = 
\left\lfloor
m_{\min} + (T - m_{\min}) \cdot 
\frac{e^{K \cdot \frac{s - \tau_{\min}}{\tau_{\max} - \tau_{\min}}} - 1}
{e^{K} - 1}
\right\rfloor ,
\quad \text{for } \tau_{\min} < s < \tau_{\max}.
\]
where $k$ determines the curvature of the mapping. A positive $k$ concentrates image conditioning into earlier diffusion steps, while a negative $k$ distributes conditioning over a longer duration. Figure~\ref{fig:timesteps_scores} illustrates the relationship between the similarity score and the corresponding timestep for different $k$ values ($-3, -1, 0, 1, 3$), showing how the exponential mapping alters the adaptive schedule.

As shown in Table~\ref{tab:exp-mapping}, the best FID is achieved by the linear mapping, indicating superior perceptual quality when the conditioning timesteps increase proportionally with similarity. Interestingly, exponential mappings reveal a clear trade-off between CLIP-I and CLIP-T. For large positive curvature ($k=3$), CLIP-T improves because image conditioning is omitted in many early steps, allowing stronger text guidance; however, this reduces CLIP-I due to weaker visual coherence. Conversely, for negative curvature ($k=-3$), CLIP-I achieves its highest value since conditioning is applied over more timesteps, thereby enhancing image-text consistency at the cost of a slightly lower CLIP-T. 

\begin{table}[t]
	\centering
	\begin{tabular}{lccc}
		\hline
		Model & CLIP-I $\uparrow$ & CLIP-T $\uparrow$ & FID $\downarrow$ \\
		\hline
		Adaptive (Linear) & 0.7608 & 0.3268 & \textbf{41.28} \\
		Exponential ($k = 1$) & 0.7603 & 0.3289 & 41.39 \\
		Exponential ($k = -1$) & 0.7618 & 0.3263 & 41.83 \\
		Exponential ($k = 3$) & 0.7611 & \textbf{0.3312} & 41.59 \\
		Exponential ($k = -3$) & \textbf{0.7633} & 0.3254 & 41.53 \\
		\hline
	\end{tabular}
	\caption{Effect of exponential mapping on adaptive timestep scheduling in AVC.}
	\label{tab:exp-mapping}
\end{table}

Overall, these results confirm that exponential scheduling provides flexible control over semantic versus visual emphasis, while the linear mapping remains the most balanced configuration in terms of fidelity and alignment.

\section{Limitations}
Although the proposed framework demonstrates consistent improvements over the baseline, several limitations remain. 
First, the backbone generative model is based on Stable Diffusion~1.5, which occasionally produces images with structural or semantic inaccuracies, particularly in complex narrative scenes. 
These imperfections are also reflected in the human evaluation results, where raters noted occasional inconsistencies between textual descriptions and visual details. 
Second, our baseline model, StoryGen, was originally trained on the StorySalon dataset; consequently, its generalization to other datasets is limited. 
Since our method builds upon this pretrained backbone rather than retraining from scratch, its overall quality on out-of-domain data inherits part of this weakness. 
Future work will explore integrating more recent diffusion architectures and retraining on diverse story visualization corpora to improve generalization and reduce visual inaccuracies.

\section{Conclusion}
In this work, we introduced \textbf{Adaptive Visual Conditioning (AVC)}, a diffusion-based framework for story continuation that dynamically adjusts the contribution of prior visual context according to its semantic alignment with the current narrative. To enhance textual supervision, we re-captioned the \textit{StorySalon} dataset using large-scale vision–language models, leading to stronger semantic consistency. We further proposed a CLIP-based semantic image selection mechanism to identify the most relevant reference frame, and an adaptive conditioning strategy that modulates the influence of visual context across diffusion timesteps. 

Extensive experiments on the \textit{StorySalon} dataset demonstrate that AVC improves narrative coherence, semantic alignment, and visual fidelity compared to strong baselines. Ablation studies further validated the effectiveness of CLIP-based selection and adaptive conditioning. Although our approach does not involve additional training, it provides a lightweight yet effective enhancement over pretrained models such as StoryGen.

 \bibliographystyle{elsarticle-harv} 
 \bibliography{all_docs/references}




% %% For authoryear reference style

% \bibitem[Lamport(1994)]{lamport94}
%   \textit{\LaTeX: a document preparation system},
%   Addison Wesley, Massachusetts,

% \end{thebibliography}
\end{document}